\documentclass[preprint, 5p, times, twocolumn]{elsarticle}

\usepackage{amssymb}
\usepackage{amsmath}

\usepackage{bookmark}
\usepackage{subfig}
\DeclareMathAlphabet{\mymathbb}{U}{BOONDOX-ds}{m}{n}
\newcommand{\tech}{\text{{MASS}}}{}
{}
\usepackage[linesnumbered,ruled]{algorithm2e}
\usepackage{threeparttable}
\usepackage{amsmath}
\usepackage{algpseudocode}
\usepackage{wrapfig}
\usepackage{needspace}
\newtheorem{Tlemma}{Lemma}

\newtheorem{Tdef}{Definition}

\usepackage{multirow}
\usepackage{xcolor}
\usepackage{subfig}

\SetCommentSty{mycommfont}    
\usepackage{pifont}
\let\oldding\ding
\renewcommand{\ding}[2][1]{\scalebox{#1}{\oldding{#2}}}

\newcommand{\minor}[1]{\textcolor{black}{#1}}
\newcommand{\major}[1]{\textcolor{black}{#1}}
\newcommand{\change}[1]{\textcolor{black}{#1}}

\journal{Journal of Systems Architecture}

\begin{document}

\begin{frontmatter}



\title{Mapping and Scheduling Spiking Neural Networks On Segmented Ladder Bus Architectures}


\author[1]{Phu Khanh Huynh}
\author[2]{Francky Catthoor}
\author[1]{Anup~Das} 

\affiliation[1]{organization={Department of Electrical and Computer Engineering, Drexel University},
            city={Philadelphia},
            postcode={19104}, 
            state={PA},
            country={United States}}
\affiliation[2]{organization={National Technical University of Athens},
            city={Athens},
            country={Greece}}

\begin{abstract}
Large-scale neuromorphic architectures consist of computing tiles that communicate spikes using a shared interconnect.
The communication patterns in such systems are inherently sparse, asynchronous, and localized due to the spiking nature of neural events, characterized by temporal sparsity with occasional bursts of traffic.
These characteristics necessitate interconnects optimized for handling high-activity bursts while consuming minimal power during idle periods. 
Dynamic segmented bus has been proposed a promising interconnect for its simplicity, scalability and low power consumption. 
However, deploying spiking neural network applications on such buses presents challenges, including substantial inter-cluster traffic, which can lead to network congestion, spike loss, and unnecessary energy expenditure.
In this paper, we propose a three-step process to deploy SNN applications on dynamic segmented buses aiming to reduce spike loss and conserve energy. 
Firstly, we formulate optimization heuristics to mitigate spike loss and energy consumption based on application connectivity. 
Secondly, we analyze the application traffic to determine spike schedules that minimize traffic flooding. 
Lastly, we propose a routing algorithm to minimize spike traffic path crossings.
We evaluate our approach using a cycle-accurate network simulator.
The simulation results show that our algorithms can eliminate spike loss while keeping energy consumption significantly lower compared to conventional NoCs.
\end{abstract}




\end{frontmatter}



\section{Introduction}\label{sec:introduction}
An event-driven neuromorphic system is a computing platform that implements biological neurons and synapses in hardware to execute spiking neural networks (SNN)~\cite{rathi2023exploring}.
To address design scalability, a neuromorphic system is designed as a many-core hardware, where cores are interconnected using a shared time-multiplexed interconnect such as shared bus, network-on-chip (NoC) (e.g., NorthPole~\cite{modha2023neural}, DYNAP-SE2~\cite{richter2024dynap}), and segmented bus (e.g., $\mu$Brain~\cite{sentryos}).

The traffic in such systems is driven by spiking events, differing fundamentally from the continuous and synchronous communication typical of traditional computing architectures. 
A defining feature of neuromorphic traffic is temporal sparsity~\cite{olshausen2004sparse, wolfe2010sparse}. Communication occurs only when a neuron generates a spike, resulting in sparse data traffic.
In large-scale neuromorphic systems, densely connected neurons are typically grouped within the same clusters, while connections between clusters remain sparse. 
This clustering leads to traffic patterns where inter-cluster communication can experience short periods of intense activity, especially when many neurons spike simultaneously, followed by long periods of inactivity.
Consequently, the communication infrastructure supporting neuromorphic system must be optimized to accommodate this characteristic: it must not only handle high-burst traffic efficiently but also consume minimal power during idle states.

Similar to other many-core architectures, SNN applications are required to be mapped to the hardware prior to execution. 
This process usually includes, but not limited to: partitioning the application into several clusters which can fit into the hardware neural core; placing the clusters into hardware based on the application connectivity; and scheduling/routing the spike traffic between cores. 
Previous works have been proposed to address mapping SNNs into neuromorphic hardware, such as SpiNeMap~\cite{spinemap} and~\cite{jin2023mapping}. 
These works show that optimized mapping can greatly influence the performance of the running SNN application, including energy consumption, latency, and inter-spike interval.

However, these methods mainly perform such mapping operations on NoCs, which is not fully suitable for running SNN applications.  
\change{The rationale for this can be attributed to the significant energy and area overhead of NoC, which is mainly due to the use of large buffers to temporarily store data packets and look-up tables (LUTs) to store routing tables inside each switch.}
While effective for conventional data handling, such design becomes resource-intensive in neuromorphic contexts where communication is sparse and irregular. 
\change{A major factor contributing to this inefficiency is the leakage power associated with the substantial memory capacity required by the buffers.}
This results in unnecessary energy consumption, which contradicts the low-power and burst-like spike traffic requirements that are critical for neuromorphic systems.
\change{To mitigate such disadvantages, bufferless NoCs have been proposed as an alternative to buffered NoCs~\cite{moscibroda2009case}.
However, in bufferless NoCs, when multiple packets compete for the same output port, all but one packet must be deflected and rerouted through alternative paths. 
This deflection process leads to additional traversals through the network~\cite{michelogiannakis2010evaluating}, thereby increasing the overall network activity and energy consumption.}

Segmented bus has been introduced as an alternative for NoC in these multi-core neuromorphic hardware systems. 
\change{While the switches in a segmented bus are also bufferless, the communication system is structured to minimize path crossings, effectively avoiding the deflection and rerouting issues typically encountered in bufferless NoCs.}
It was shown that segmented buses such as NeuSB~\cite{balaji2022NeuSB} outperforms NoCs in both latency and energy consumption for running SNN applications. 
Nevertheless, NeuSB lacks runtime switch controls and thus does not truly support running different applications on the same hardware. 
In this work, we focus on segmented ladder bus~\cite{huynh2024adiona}, the latest dynamic segmented bus interconnect that can offer runtime configuration for routing and scheduling network traffic.

We propose a mapping/scheduling framework, namely \textbf{\tech{}} (\textbf{MA}pping and \textbf{S}cheduling \textbf{S}NNs), that can fully deploy any SNN to segmented ladder bus architectures. 
Our main contributions for this paper are as follows:
\begin{itemize}
    \item We propose a mapping algorithm based on Hill Climbing technique that can help minimize both energy consumption and spike loss in the network. 
    \item We are the first to show a spike traffic scheduling and routing algorithms in dynamic segmented ladder bus.
    \item \change{We evaluate our algorithms individually and combine to show the contribution of each step and prove that overall segmented ladder bus can indeed support large SNN applications while preserving energy consumption.}
\end{itemize}

The remainder of the paper is organized as follows. A background introduction regarding dynamic segmented ladder bus and related mapping works are presented in Section~\ref{sec:background}. 
The overview of the proposed solutions and detailed algorithms are presented in Section~\ref{sec:solution}. 
Evaluation methodology is presented in Section~\ref{sec:evaluation}. 
Results and discussions are presented in Section~\ref{sec:results}. 
Finally, the paper is concluded in Section~\ref{sec:conclusions}.

\section{Background and Related Works}\label{sec:background}
This section presents the requisite background to comprehend the segmented ladder bus~\cite{huynh2024adiona}, the primary interconnect architecture targeted by our mapping and scheduling methodologies.
Furthermore, we review related works on the mapping of SNNs to neuromorphic hardware architectures, emphasizing the associated challenges and recent innovations in this domain.
\subsection{Segmented Ladder Bus}\label{sec:ladder_bus_basic}
The architecture of a segmented ladder bus is designed with the following characteristics:
\begin{itemize}
    \item The tiles are logically divided into two parallel rows, each comprising an equal number of tiles.
    \item A fixed number of parallel segmented bus lanes are placed between these tile rows.
    \item The connections between the tiles and parallel bus lanes are established using criss-cross three-way switches.
\end{itemize}

A segmented ladder bus is governed globally by a compile time software framework and locally use coarse-grain runtime hardware controllers in order to enable dynamic configuration of switches.
These local controllers encompass predefined switch change scenarios and facilitate the transmission of control signals to the switches accordingly.
By issuing the control signals from the controllers with the correct timing, the switches can be used to create paths that support multiple simultaneous connections inside the network, matching application requirements.
The controllers are implemented in a distributed manner, where each controller is responsible for managing the switches within its vicinity, similar to a distributed loop controller~\cite{raghavan2008distributed}. 
As a result, in a large-scale network, numerous controllers will be deployed to effectively govern the switches throughout the system.
\begin{figure}[h!]
    \centering
    \centerline{\includegraphics[width=1\columnwidth]{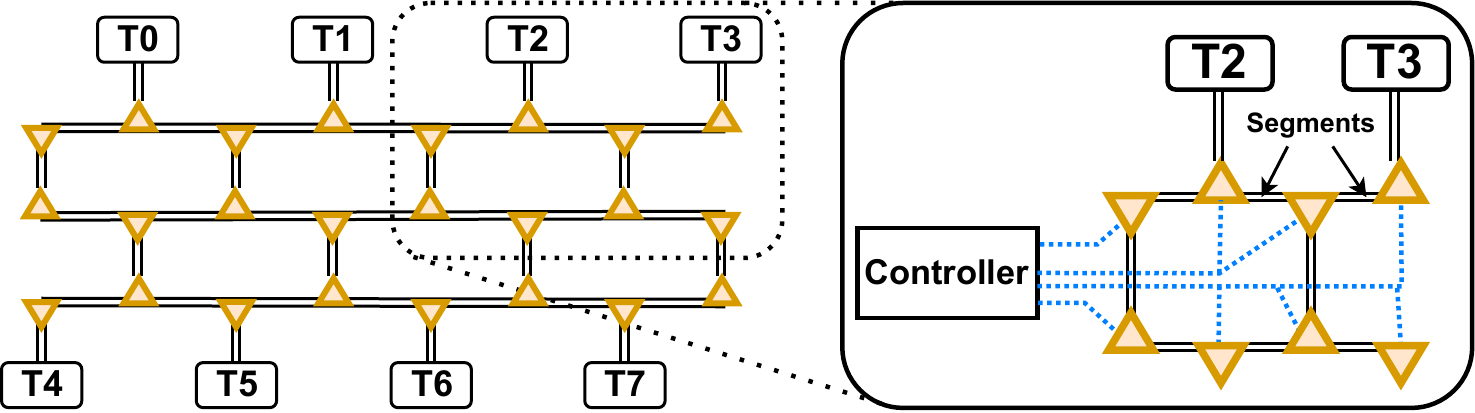}}
    \caption{An example of segmented ladder bus with a switch controller.}
    \label{fig:ladder_bus_example}
    \vspace{-10pt}
\end{figure}

Figure~\ref{fig:ladder_bus_example} illustrates an exemplary implementation of a segmented ladder bus comprising 8 tiles and 3 parallel segmented bus lanes. 
It is important to highlight that both the number of tiles and bus lanes can be scaled according to the specific requirements of the target application.
This design framework offers several key advantages:
\begin{itemize}
    \item \textbf{High Flexibility}: The network architecture facilitates communication between any pair of tiles within the system, ensuring robust interconnectivity.
    \item \textbf{Enhanced Routing and Fault-Tolerance}: Multiple distinct paths are available for communication between any two tiles, providing diverse routing options and improving the system's fault-tolerance capabilities. 
    \item \textbf{Run-Time Configurability}: The switches can be configured during run time, allowing the segmented ladder bus to provide scheduling options under network congestion. 
    \item \textbf{Energy and Area Efficiency}: The bufferless data plane significantly reduces energy consumption and minimizes the interconnect design area.
\end{itemize}
\major{Compare to NoCs, the segmented ladder bus sacrifices general-purpose routing flexibility in exchange for compile-time optimized, energy-efficient communication tailored to the sparse and bursty traffic spiking trains in neuromorphic workloads. While NoCs offer higher runtime flexibility and better fault tolerance due to their adaptive routing and buffering, they also incur significant static and dynamic energy overheads and complexity, which are often unnecessary for SNNs. The ladder bus, being bufferless and pre-scheduled, can operate at lower power and area cost while meeting the timing and bandwidth requirements of neuromorphic applications. Fault tolerance in our approach can be addressed at the compile-time mapping level by reassigning communication paths around faulty switches or tiles.}

\major{Although the segmented ladder bus is bufferless and lacks intermediate synchronization points, this does not hinder its suitability for neuromorphic systems. In this domain, precise timing closure is not a strict requirement, as these systems are event-driven and tolerant to timing variations~\cite{serrano2013stdp}. As long as spike event ordering and causality are preserved, variations in communication latency across longer paths do not compromise application correctness. This is because spike trains typically have inter-spike timing distances in the order of milliseconds or higher. Therefore, while signal delays may vary slightly across long paths, potentially reaching {\textmu}second levels, the impact on application level correctness is negligible due to the above-mentioned millisecond-scale timing requirements of neuromorphic workloads. This temporal flexibility makes the segmented ladder bus particularly well-suited for energy-efficient, pre-scheduled interconnects that do not require cycle-level synchronization across the chip.}

\begin{figure}[h!]
    \centering
    \centerline{\includegraphics[width=1\columnwidth]{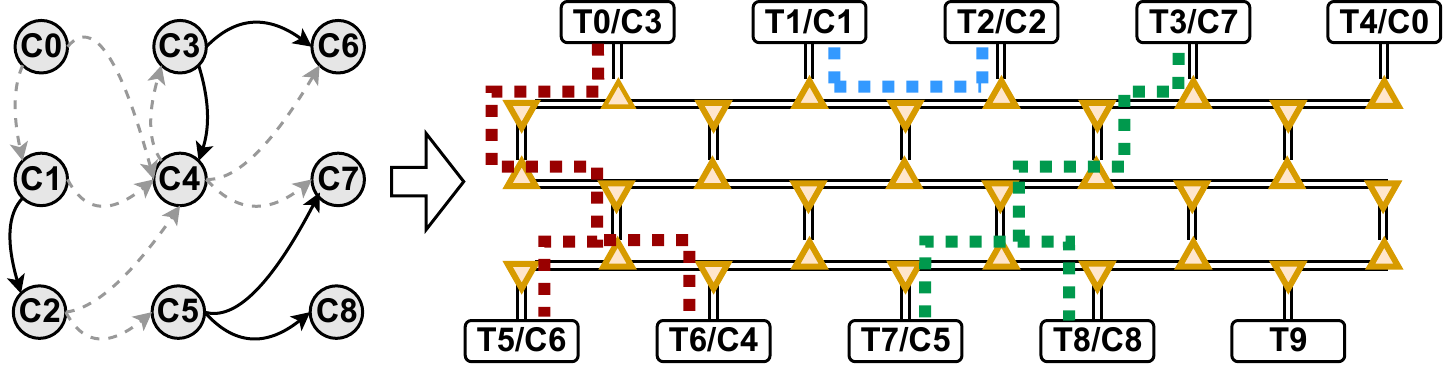}}
    \caption{Multiple routing options in a segmented ladder bus}
    \label{fig:ladder_bus_advantages}
    \vspace{-10pt}
\end{figure}

A more comprehensive understanding of the advantages offered by segmented ladder bus can be gained through an examination of Figure~\ref{fig:ladder_bus_advantages}. 
\major{The cluster communication graph is presented on the left with each cluster denoted by a unique identifier from C0 to C8.
In order to be deployed for execution, each cluster is sized so that it fits on a single hardware tile (neuromorphic core).
With a suitable mapping (e.g., C0 to T4, C1 to T1, etc.),} multiple communication paths can be facilitated simultaneously: Tile 0 sends spikes to Tiles 5 and 6, Tile 1 sends to Tile 2, and Tile 7 to Tiles 3 and 8.

\subsection{Related Works}\label{sec:related_works}
Mapping neuromorphic clusters to hardware cores in network-on-chip (NoC) architectures has garnered significant attention in recent years due to its potential to enhance the performance and efficiency of neuromorphic computing systems. Several notable approaches and frameworks have been proposed to address this challenge.

One prominent method is SpiNeMap~\cite{spinemap}, which focuses on optimizing the mapping of spiking neural networks (SNNs) onto neuromorphic hardware. 
\major{SpiNeMap uses multi-objective optimization to balance energy, latency, and inter-spike interval based on SNN-specific communication patterns.
Similarly, SNEAP~\cite{li2020sneap} employs heuristics such as Simulated Annealing, Particle Swarm Optimization, and Tabu Search, but its dependence on static NoC topologies limits scalability and adaptability under dynamic workloads.}

\major{Jin et al.~\cite{jin2023mapping} propose a heuristic approach combining Hilbert curve-based placement with a Force Directed refinement, leveraging SNN sparsity to reduce communication overhead. Despite its effectiveness, the heuristic nature of this algorithm may not always guarantee optimal mappings, especially in larger and more complex networks.}



\major{NeuMap~\cite{neumap} introduces a machine learning-based mapper that adapts to workload variations using predictions from trained models. Though promising in adaptability, it incurs training overhead and depends on extensive historical data, which may hinder real-time deployment.}

In summary, the field of mapping neuromorphic clusters onto interconnect architectures is abundant with innovative approaches and significant advancements. 
However, existing approaches primarily target NoCs and fail to address the unique challenges associated with mapping applications onto segmented bus-based neuromorphic hardware.
Therefore, there is a significant need for more tailored and efficient techniques for mapping, scheduling, and routing neuromorphic clusters on segmented ladder bus architectures.

\section{Proposed Solutions}\label{sec:solution}
In our previous work~\cite{mustafazade2024clustering}, we have described the SNN application partitioning process in more details. 
This allocates neurons to memristive crossbar clusters while maximizing the crossbar utilization and minimizing inter-cluster communication.
Although such process is essential, it works independently of the types of interconnect hardware. 
\major{In particular, the algorithms take only the spike traffic information from the application as input and do not account for the architectural constraints or design opportunities introduced by the segmented ladder bus or other specific interconnects.}
For this reason, it is not the focus of this paper.

\begin{figure}[h!]
	\centering
	\centerline{\includegraphics[width=1\columnwidth]{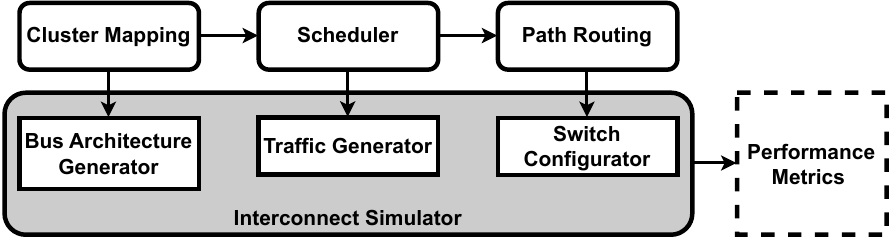}}
	\caption{\tech{} design flow.}
	\label{fig:design_flow}
\end{figure}

Once an SNN application has been partitioned into clusters, we need to: map the clusters to actual hardware tiles; schedule spike traffic; and route the scheduled traffic. 
Each of these steps can be optimized to improve the hardware utilization of the network, guarantee performance of the application, and reduce energy consumption.
Figure~\ref{fig:design_flow} shows our proposed solution processes integrated inside our evaluation framework.
\change{
While numerous state-of-the-art network simulators provide support for Network on Chip architectures, none currently offer built-in capabilities for simulating segmented bus architectures.
Thus, we developed a custom network simulator tailored to segmented busesbased on Noxim~\cite{catania2015noxim}.
Noxim was chosen as the foundation for our simulator due to its open-source, extensible codebase, SystemC-based~\cite{systemC} cycle-accurate simulation engine, and lightweight, scalable design.
}

\subsection{Cluster Mapping}\label{sec:cluster_map}
After an SNN application has been partitioned into clusters, it is important to map these clusters into actual hardware tiles. The connectivity between the clusters can be represented as a directed graph ${G_\mathit{CSNN} = (C, L)}$ consisting of $C$ clusters and $L$ links between the clusters. 
A link ${L_{i,j}\in L}$ is a channel connecting cluster $C_i$ with cluster $C_j$, where $C_i$, $C_j \in C$. 
Each link $L_{i,j}$ has a weight $s_{i,j}$ that represents the number of spikes communicated on $L_{i,j}$.
A spike link activity indicator $z_{i,j} = 1$ if $s_{i,j} > 0$ and $s_{i,j} = 0$ otherwise. 
The cluster mapping problem can be modeled as an injective function $\phi$ that place the neuron clusters $C$ into hardware tiles $T$. 

\begin{figure}[h!]
    \centering
    \captionsetup[subfloat]{labelfont=normalsize,textfont=normalsize}
    \subfloat[Topographically crossed paths\label{fig:topo_crossed}]{{\includegraphics[width=0.48\columnwidth]{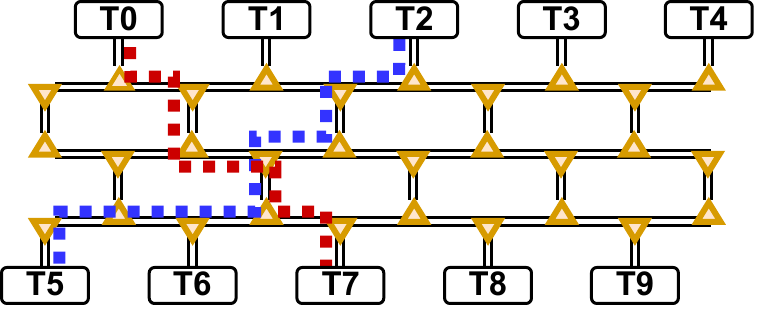}}}
    \hfill
    \subfloat[Topographically non crossed paths \label{fig:topo_no_crossed}]{{\includegraphics[width=0.48\columnwidth]{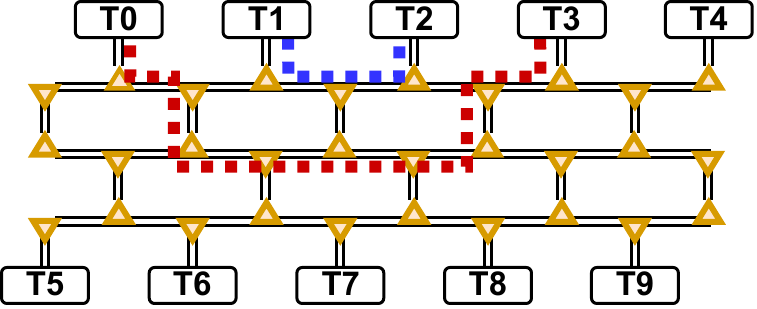}}}
    \caption{Example of topographically crossing.}
    \label{fig:topo_example}
    \vspace{-10pt}
\end{figure}

Different mappings will have different impacts on energy consumption, latency, and the number of crossed communication paths. 
In our segmented ladder bus design, communication energy consumption is proportional to the length of the communication paths and the number of switch configurations that need to be changed.
Furthermore, unlike NoCs, the bufferless nature of the switches in segmented bus causes simultaneous intersecting communication paths to result in either packet drops or additional delays at the source nodes to avoid packet loss.
Thus, we propose two different mapping cost functions specifically adapted to ladder segmented bus to minimize either dynamic energy consumption or number of crossed paths. 
\major{We focus on dynamic energy because, for any given application, the same hardware is used across all mapping, scheduling, and routing configurations. As a result, static (leakage) power remains largely constant and does not impact the relative comparison between configurations. Including static power would only introduce a uniform offset to the total energy values. Therefore, dynamic energy more accurately reflects the impact of interconnect activity and the effectiveness of different mapping strategies.}
They are computed as follows.
\begin{enumerate}
    \item \textit{Dynamic Energy:}
    \begin{equation}
    \label{eq:cluster_energy}
    E = \sum_{i, j \in \{0,1,\cdots,|C|\}}s_{i,j}\times d_{\phi(i), \phi(j)}
    \end{equation}
    where $d_{\phi(i), \phi(j)}$ is the number of segments between the tiles that $C_i$, $C_j$ are mapped to.  
    \item \textit{Weighted crossed paths:} Since our interconnect architecture is completely bufferless, we need to find the maximum number of simultaneous connections to prevent spike loss and minimize the number of switching in the network. \change{Although ladder segmented bus can support multicasting, it is important to note that this work focuses on unicast communication, and the extension to multicast scenarios is left for future work.}
    \begin{multline}
    \label{eq:crossed_paths}
    W = \sum_{i,j,k,l \in \{1,\cdots,C\}}(s_{i,j} + s_{k,l})\times z_{i,j}\times z_{k,l} \\ \times crossed(\phi(L_{i,j}), \phi(L_{k,l}))
    \end{multline}
    where $crossed(\phi(L_m), \phi(L_n))$ is the function to check whether a mapping would make the links $L_m$ and $L_n$ topographically cross each other (see Figure~\ref{fig:topo_example}).
\end{enumerate}
One can then select or combine the desired cost function depends on the application requirements, expressed as follows:
\begin{equation}
    \label{eq:cost_mapping}
    Cost = \alpha \times E + \beta \times W
\end{equation}
where $\alpha$ and $\beta$ are adjustable parameters.

Given that cluster mapping is a well-established NP-hard problem~\cite{gary1979computers}, a practical solution is implemented using a Hill Climbing approach~\cite{selman2006hill}.
This metaheuristic provides an efficient and practical way to explored the vast solution space while balancing computational cost and solution quality.
We employ a specific variant, called steepest ascent Hill Climbing, to enhance the search process.
The algorithm starts with a random mapping solution and compares the cost of that solution with all neighboring solutions. 
A neighbor is defined as a mapping obtained by swapping the positions of two nodes in the current mapping.
This ensures that the search space of the algorithm is limited to only valid mappings, which improve the efficiency of the search.
The algorithm moves the solution to the neighbor with the most improved cost. The search continues until there is no neighbor with improved cost.
The local search ends here, with the algorithm keeping record of this solution. 
It is important to note that Hill Climbing may converge to a local minima.
To mitigate this, when the search terminates due to the lack of better-cost neighbors, we incorporate a number of random swaps and re-initiate the search procedure. 
Ultimately, the algorithm concludes its execution after running a predetermined number of random perturbations.
\begin{algorithm}[h!]
	\scriptsize{
        \KwIn{$G_\mathit{CSNN}$, $\eta$, $I$}
        \KwOut{Mapping $\phi$: $C \rightarrow T$}
        $\phi_{best} = random\_map(C, T)$\;
        \For(\tcc*[f]{loop over the number of perturbations}){$i = 1$ to $\eta$}{
            $\phi_{current} = random\_map(C, T);$ \tcc*[f]{generate random mapping}\\
            $\phi_{i} = \phi_{current}$;\tcc*[f]{set best local mapping to the current mapping}\\
            $improved = \textbf{true}$\;
            $k = 0$\;
            \While(\tcc*[f]{while cost is improved and under the number of iteration limit}){$improved$ \textbf{and} $k < I$}{
                $k = k + 1$\;
                $improved = \textbf{false}$\;
                $change_{best} = +\infty$\;
                \For(\tcc*[f]{iterate over all neighbours}){$\phi_n \in \phi_{current}.neighbours()$}{
                    $change = cost(\phi_{n}) - cost(\phi_{current})$;\tcc*[f]{compare the cost of current mapping and its neighbour}\\
                    \If(\tcc*[f]{if the change of cost is lower than the best change}){$change < change_{best}$} { 
                       $change_{best} = change$\;
                       $\phi_{i} = \phi_n$;\tcc*[f]{set best local mapping to the neighbour mapping}\\
                    }
                }
                \If{$change_{best} < 0$} {
                    $\phi_{current} = \phi_{i}$;\tcc*[f]{move current mapping to the best neighbour cost mapping}\\
                    $improved = \textbf{true}$;\tcc*[f]{set improvement flag if cost change in the right direction}\\
                }
            }
            \If (\tcc*[f]{if cost of best local mapping is lower}){$cost(\phi_{i}) < cost(\phi_{best})$}{
               $\phi_{best} = \phi_{i}$;\tcc*[f]{set best global mapping to best local mapping}\\
            }
        }
        \textbf{return} $\phi_{best}$
    }
    \caption{{Cluster mapping algorithm.}}
    \label{alg:cluster_mapping}
\end{algorithm}


\textbf{Time Complexity Analysis:} The time complexity of Algorithm~\ref{alg:cluster_mapping} is computed as follows. A mapping of $C$ clusters to $T$ hardware nodes has a total of $T^2$ neighbours.
For each local search, the number of iteration is limited to $I$, which in our experiments is set proportional to the number of tiles $T$. 
If the maximum allowed number of perturbation is $\eta$ the overall time complexity is: $O(\eta T^3)\cdot O(cost)$. 
\begin{itemize}
    \item The weighted crossed path cost function evaluates crossings between every pair of links. With $L$ links, this results in $L^2$ crossings, leading to a time complexity of $O(\eta T^3 L^2)$.
    \item For the dynamic energy cost function, it calculates the weighted sum across all $L$ links. The resulting time complexity is $O(\eta T^3 L)$.
\end{itemize}
This polynomial complexity is manageable, as the mapping is performed only once at compile time.

\subsection{Spike Traffic Scheduling}\label{sec:spike_scheduling}
Despite efforts to minimize crossings between communication paths through our cluster mapping algorithm, the high density and frequency of application connectivity can occasionally prevent simultaneous facilitation of all required communication links. 
If the number of required simultaneous communication links exceeds the network's capacity, it can result in spike loss due to the bufferless nature of our architecture.
Hence, it becomes necessary to schedule spike traffic to maintain application performance.

With the cluster mapping acquired from algorithm~\ref{alg:cluster_mapping} and the spike traffic from CARLSim~\cite{carlsim6} or SNNTorch~\cite{snntorch}, we can obtain the exact timing of the spike traffic among the hardware tiles.
For any given time with multiple simultaneous communication links, the scheduler would need to find groups of links for which no two links in any group intersect, while minimizing the number of groups to prevent unnecessary delays.

\begin{figure}[h!]
    \centering
    \captionsetup[subfloat]{labelfont=normalsize,textfont=normalsize}
    \subfloat[Full simultaneous traffic\label{fig:full_traffic}]{{\includegraphics[width=0.5\columnwidth]{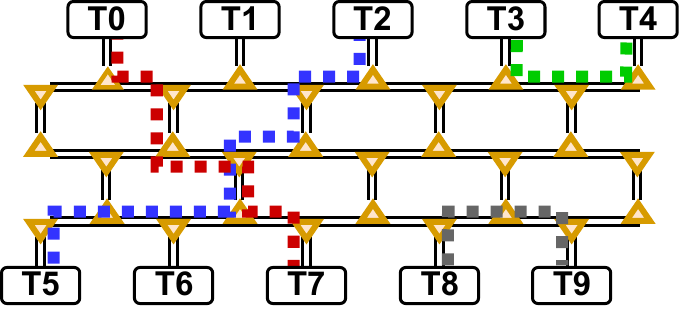}}}
    \hfill
    \subfloat[Group all possible non-crossing paths and then schedule the remaining ones.  \label{fig:grouped_traffic}]{{\includegraphics[width=1\columnwidth]{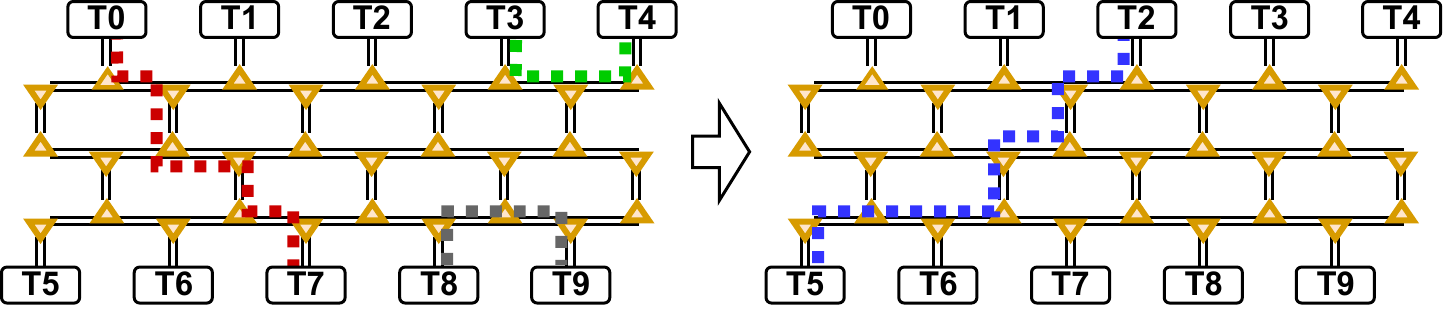}}}
    \caption{Example of scheduling traffic.}
    \label{fig:scheduling_example}
    \vspace{-10pt}
\end{figure}

This problem is essentially a bin packing problem with conflicts (BPPC)~\cite{ekici2021bin}, which is NP-complete. 
Therefore, instead of seeking an optimal schedule, we develop a custom heuristic to get a solution within a practical computational limit. 
An example of a spike traffic schedule can be found in Figure~\ref{fig:scheduling_example}.

\textbf{Heuristic Solution:}
The exact timing of the spike traffic among hardware tiles is denoted as $T_{O}$.
We then iterate over all the traffic time steps in $T_{O}$.
When multiple communication links need to run at the same time step, they are sorted according to the number of spikes communicated. 
We subsequently go through the sorted list, and attempt to group links that can operate concurrently without their paths intersect. 
Links in the same group can be scheduled to run at the same time.
If a link intersects with any other link, a new group is formed with an additional delay for that group.
This intersection check is based on the graph representation of the ladder bus $G_{ladder}$.
Note that this check can be done using topological crossing or shortest path crossing of links.
\change{The utilization of bus lanes is also taken into account when grouping the communication links.
If adding a new link exceeds the available bus lane capacity, it is treated as intersecting with other links in the group.}
This process continues until all links are grouped with specific run times, ensuring that there are no intersections within any group.
This scheduling process can be performed without affecting the application performance because the speed of the bus is much faster than the communication frequency of the neuron spikes.

\begin{algorithm}[h!]
    \scriptsize{
    \KwIn{$G_{ladder}$, $T_O$}
    \KwOut{$Scheduler$}
    $Scheduler = \emptyset$\;
    \For(\tcc*[f]{iterate over all traffic time steps}){$time\_step \in T_O$}{
        $SortedLinks = sort(T_O[time\_step])$;\tcc*[f]{sort the links in each time step}\\
        $TimeGroups = \emptyset$;\tcc*[f]{create a list of groups for the current time step}\\
        \For(\tcc*[f]{go through all the links in the sorted link list}){$L \in SortedLinks$}{
            $added = False$\;
            \For(\tcc*[f]{for each group in the current time step}){$Group \in TimeGroups$}{
                $crossed = False$\;
                \For(\tcc*[f]{for each link in the group}){$L_G \in Group$}{
                    \If(\tcc*[f]{check whether the current link L crosses with $L_G$}){$check\_cross(G_{ladder}, L, L_G)$} {
                        $crossed = True$\;
                    }
                }
                \If(\tcc*[f]{if link L does not cross any link in the group}){$crossed == False$} {
                    $Group.append(L)$;\tcc*[f]{add link L to the group}\\
                    $added = True$;\tcc*[f]{mark link L as added}\\
                }
            }
            \If(\tcc*[f]{if link L has not been added to any group}){$added = False$} {
                $TimeGroups.append(Set(L))$;\tcc*[f]{create a new group for L and add it to the list of groups}\\
            }
        }
        $delay = 0$\;
        \For(\tcc*[f]{for each group in the current time step}){$Group \in TimeGroups$}{
            $max\_spike = 0$\;
            \For(\tcc*[f]{for each link in the group}){$L_G \in Group$}{
                $L_G.time = time\_step + delay$;\tcc*[f]{increase scheduled time of the link by the delay amount}\\
                \If{$max\_spike < L_G.spikes$} {
                    $max\_spike = L_G.spikes$; \tcc*[f]{find the maximum spikes sent in the group}\\
                }
            }
            $delay = delay + (max\_spike \times N$);\tcc*[f]{increase the delay proportionally to the maximum spikes sent}
        }
        $Scheduler.append(TimeGroups)$;\tcc*[f]{add the list of groups to the scheduler}\\
    }
    \textbf{return} $Scheduler$
    }
    \caption{{Spike traffic scheduling algorithm.}}
    \label{alg:scheduling}
\end{algorithm}

\textbf{Time Complexity Analysis:}
\begin{itemize}
    \item For each time step, the links are sorted. Complexity of sorting $L$ simultaneous links: $O(L \log L)$.
    \item For each link $l$, the algorithm checks existing groups to see if the link can be added without conflict. For each group, the algorithm checks if the new link crosses with any existing link in the group. This involves iterating over each link in the group, and for each pair of links, the \texttt{check\_cross} function is called, which can be simplified to take constant time $O(1)$ due to the defined and regular topology of the network. In the worst case, each link will be compared to all other $L-1$ links, giving $O(L^2)$ comparisons.
    \item The delay adding steps from line 23 to line 33 just passes through the full list of links, thus has $O(L)$ running time.
    \item Hence, the scheduling complexity per time step is $O(L^2)$.
\end{itemize}

\subsection{Path Routing}\label{sec:path_routing}
Once the scheduling process has been completed, the next crucial step is to perform routing of all the simultaneous connections. 
Efficient routing is essential to optimize the interconnect performance, and reduce unnecessary delays while still maintaining the maximal number of simultaneous connections.

Our solution to the path routing problem primarily adopts a greedy algorithm, supplemented by A*~\cite{duchovn2014path} sub-processes for efficient shortest path finding. 
\change{We sort all the simultaneous communication links $CLinks$ according to the following order: we prioritize all the links with source and destination tiles on the same bus lane first, followed by sorting based on the orientation of the links, and finally by the number of spikes on each link.}
The reasoning of this sort is three folds: 
Firstly, we reduce the number of potential crossing between links by routing the links with source and destination tiles on the same bus lane and the shorter distance ones first. 
This is because the shortest path for tiles on the same bus lane is unique, while there can be many paths with the same length for tiles on different bus lanes.
\change{Secondly, links with the same orientation can be easily routed one by one so that they do not cross each other.}
Thirdly, by prioritizing to route the links with more spikes, the paths for these links will be shorter, which means lowering the number of spikes that need to travel long distances, which in turn helps in reducing the energy consumption in the network. 
We then represent the segmented ladder bus interconnect as an undirected graph $G_{ladder}$ with tiles and switches as vertices and bus wires as edges. 
The initial weights of all vertices and edges in the graph are set to 1.
The algorithm processes the sorted communication links sequentially, and uses a modified A* algorithm to determine the shortest weighted path for each specific link.
After determining the shortest path for the selected link, the weights of all vertices are increased proportionally to the number of spikes communicated on this particular link. 
Subsequently, the routed link is removed from the list of communication links.
The algorithm continues the process iteratively, selecting and routing the most communicated links one by one, until all paths are effectively routed.
\begin{algorithm}[h!]
    \scriptsize{
    \KwIn{$G_{ladder}$, $CLinks$}
    \KwOut{$RoutedList$}
    $SLinks = sort(CLinks)$;\tcc*[f]{sort the communication links}\\
    $RoutedList = \emptyset$\;
    \For(\tcc*[f]{each link in the sorted link list}){$L_{i,j} \in SLinks$}{
        $Path = A\_star(i, j)$; \tcc*[f]{find the path for the link using A*}\\
        \For(\tcc*[f]{for every node in the path}){$node \in Path$}{
            $node.weight = node.weight + L_{i,j}.spikes$;\tcc*[f]{increase the weight of the node by the number of spikes on the link}\\
        }
        $RoutedList.add(Path)$;\tcc*[f]{add the path to the routed list}\\
    }
    \textbf{return} $RoutedList$
    }
    \caption{{Path routing algorithm.}}
    \label{alg:path_routing}
\end{algorithm}

\textbf{Time Complexity Analysis:} The time complexity of Algorithm~\ref{alg:path_routing} is computed as follows. 
Sorting $L$ communication links takes $O(L \log L)$ time. 
In worst case scenario, A* algorithm takes $O(|E|+|V|)$ time where $E$ and $V$ are both proportional to the number of tiles $T$ and bus lanes $n$ in our architecture. We need to perform A* for $L$ links. Thus, the overall time complexity is $O(nLT)$.

\section{Evaluation Methodology}\label{sec:evaluation}
\subsection{Benchmarks}\label{sec:benchmarks}
\major{The hardware implementation details of the segmented ladder bus were presented in our prior work~\cite{huynh2024adiona}. In that work, we implemented both the 3-way segmented switches and a standard NoC router on an FPGA to evaluate their latency and power characteristics.
These measured values were then used to calibrate our cycle-accurate simulator extended from Noxim~\cite{catania2015noxim}, enabling it to model the segmented ladder bus alongside NoC variations with realistic energy and performance metrics.
All evaluations are based on this extended simulator, which incorporates FPGA-measured dynamic energy and latency per hop and per switch configuration.} 

Table~\ref{tab:apps} reports the machine learning applications utilized in the assessment of mapping SNNs into \tech{}.
In each instance, a pruning methodology is employed to eliminate weights close to zero. 
For each model, we provide detailed information on the total number of clusters generated to run the application.
\change{Empirically, the number of bus lanes allocated per application is set to approximately the square root of the total number of clusters.}
\begin{table}[!h]
    \renewcommand{\arraystretch}{1.0}
    \centering
    \resizebox{\columnwidth}{!}{%
	\begin{tabular}{c|c|c|c|c}
		\hline
		\textbf{Application} & \textbf{\# Clusters} & \textbf{Avg. Degree} & \textbf{Network Density} & \textbf{\# Bus Lanes} \\
        \hline
		mnist & 12 & 1.50 & 0.27 & 4 \\
        LeNet & 14 & 2.93 & 0.45 & 4 \\
  		fashion-mnist~\cite{xiao2017fashion} & 24 & 5.33 & 0.46 & 5 \\
		cifar10 & 25 & 5.44 & 0.45 & 5 \\
    	emnist~\cite{cohen2017emnist} & 30 & 5.37 & 0.37 & 6 \\
		ResNet~\cite{he2016deep} & 96 & 11.13 & 0.23 & 10 \\
		\hline
	\end{tabular}}
    \caption{\change{Applications used to evaluate our mapping and scheduling approaches.}}
    \label{tab:apps}
    \vspace{-10pt}
\end{table}

\subsection{Evaluation Approaches}\label{sec:eval_approach}
We evaluate the following approaches.
\begin{itemize}
    \item \textbf{ADIONA} or \textbf{\tech{}-SL} (Spike Loss): This baseline approach, previously used in ADIONA, focuses on mapping with the goal of reducing spike loss, without scheduling, and utilizes simple shortest path routing.
    \item \textbf{\tech{}-DE} (Dynamic Energy): This approach emphasizes mapping to minimize dynamic energy consumption, without scheduling, and also uses simple shortest path routing.
    \item \textbf{\tech{}-TXS} (Topological Crossing Schedule): This approach builds on \tech{}-SL by adding scheduling with a topological crossing check, while still applying simple shortest path routing.
    \item \textbf{\tech{}-SPXS} (Shortest Path Crossing Schedule): This approach is based on \tech{}-SL, incorporating scheduling with a shortest path crossing check, and also employs shortest path routing.
    \item \textbf{\tech{}-SR} (Sorted Routing): This approach is based on \tech{}-TXS with scheduling using topological crossing check, and uses our custom routing algorithm with sorted communication links.
\end{itemize}

\section{Results and Discussions}\label{sec:results}


\subsection{Cluster mapping}\label{sec:mapping_results}
\begin{figure}[h!]
	\centering
	\centerline{\includegraphics[width=1\columnwidth]{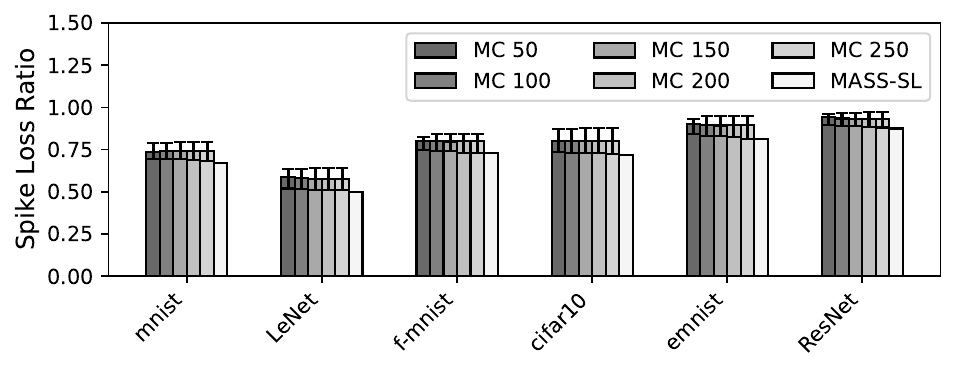}}
	\caption{Spike loss ratio for different mappings.}
	\label{fig:spike_loss_mapping}
\end{figure}

In this experiment, we evaluate our mapping algorithm for optimizing both spike losses and energy consumption against random mappings obtained from 50 to 250 Monte Carlo simulations. 
Figure~\ref{fig:spike_loss_mapping} plots the spike loss ratio for different mappings on the benchmark applications.

\major{Reducing spike loss is critical in spiking neural networks, as each spike encodes meaningful temporal information that contributes to downstream neuronal activations. Lost spikes, particularly in later or sparsely connected layers, can lead to insufficient membrane potential buildup, delayed firing, or complete suppression of neuron activity, which may ultimately result in degraded inference accuracy. By reducing spike loss during communication, our framework ensures that application-level functionality and accuracy are preserved.}

On average, spike loss using our mapping algorithm and cost function is reduced by 10\% compared to the average values founded using Monte Carlo method. 
Our mapping has lower spike loss percentage than the minimum values founded using Monte Carlo method. 

\begin{figure}[h!]
	\centering
	\centerline{\includegraphics[width=1\columnwidth]{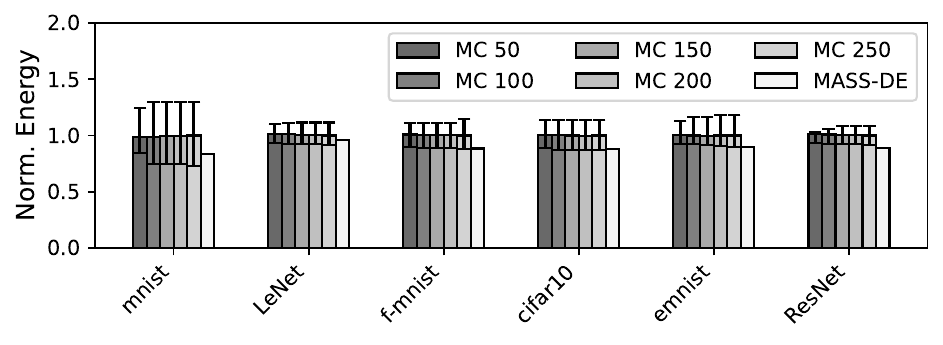}}
	\caption{Interconnect energy for different mappings.}
	\label{fig:energy_random_map}
\end{figure}

Figure~\ref{fig:energy_random_map} plots the energy consumption between a random mapping and the optimized energy mapping running on the benchmark applications.

On average, energy consumption using our mapping algorithm and cost function is reduced by 13.5\% compared to the average values founded using Monte Carlo method. 
However, the minimum energy consumption founded using Monte Carlo method for mnist application is still lower than the value founded using our algorithm. 
The reason is because our energy cost function does not take into account spike loss on the network. 
When spike packets are dropped, the network does not consume dynamic energy for transporting those packets anymore, thus the energy consumption can be lower compared to theoretical cost calculation. 
\begin{table}[h!]
    \centering
    {\fontsize{8.5}{8.5}\selectfont
    \begin{tabular}{l|c|c|c}
    \hline
    \textbf{Models} & \textbf{Monte Carlo} & \textbf{\tech{}-SL} & \textbf{\tech{}-DE} \\
    \cline{1-4}
    mnist   & 5h 28m 29s & 1m 33s  & 1m 20s  \\
    LeNet   & 8h 45m 12s & 3m 10s & 1m 57s \\
    f-mnist & 15h 55m 03s & 38m 26s & 3m 44s   \\
    cifar10 & 20h 38m 26s & 47m 23s & 4m 36s \\
    emnist  & 22h 12m 50s & 1h 03m 39s & 5m 30s \\
    ResNet  & 33h 20m 45s & 10h 52m 29s & 13m 46s \\
    \hline
    \end{tabular}}
    \label{tab:running_time_mapping}
    \caption{Running time for different mappings.}
\end{table}

Our algorithm executes significantly faster than the Monte Carlo method when using either the spike loss or energy cost function. 
Energy mapping outpaces spike loss mapping because the spike loss cost function is more complex compared to the energy cost function.

\subsection{Scheduling and Routing}\label{sec:scheduling_results}
\begin{figure}[h!]
	\centering
	\centerline{\includegraphics[width=1\columnwidth]{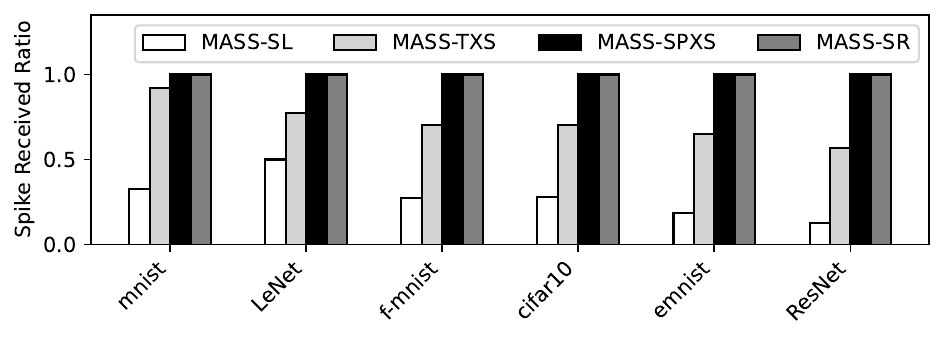}}
	\caption{Spike received ratio for non scheduled and scheduled traffic.}
	\label{fig:spike_received_schedule}
\end{figure}

\begin{figure}[h!]
	\centering
	\centerline{\includegraphics[width=1\columnwidth]{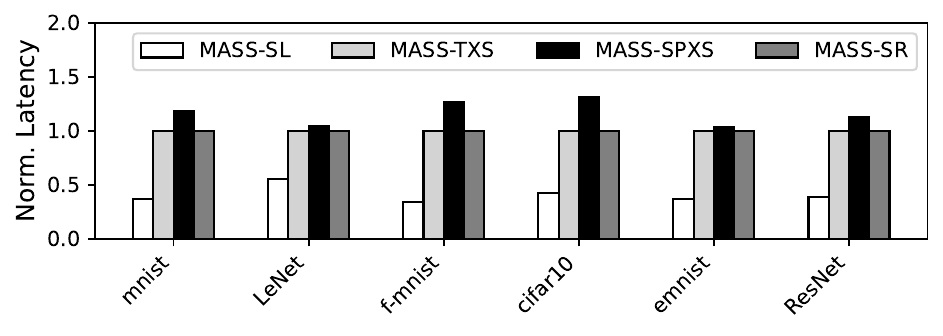}}
	\caption{Normalized latency for non scheduled and scheduled traffic.}
	\label{fig:latency_schedule}
\end{figure}

We compare the trade-off between spike loss and latency for different scheduling options. 
The scheduling algorithms used here are: \tech{}-SL (no scheduling), \tech{}-TXS, \tech{}-SPXS, and \tech{}-SR.
Figure~\ref{fig:spike_received_schedule} plots the spike received ratio (the complimentary number of spike loss ratio) and Figure~\ref{fig:latency_schedule} plots the normalized latency for scheduled and non-scheduled traffic running on the benchmark applications.

\begin{figure}[h!]
	\centering
	\centerline{\includegraphics[width=1\columnwidth]{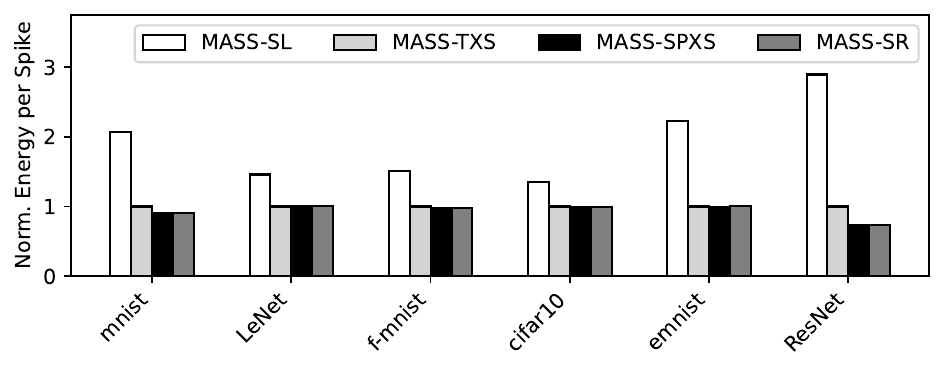}}
	\caption{Normalized energy per spike for unscheduled and scheduled traffic.}
	\label{fig:eps_overall}
    \vspace{-10pt}
\end{figure}

From these figures, we can clearly see the trade-offs between spike loss and latency using different scheduling options. 
With no scheduling, the spike loss level is very high at more than 50\% while latency remains low. 
If we spread out the traffic more, the spike loss level can become lower until we achieve zero spike loss with \tech{}-SPXS.
With less spike loss, the energy per spike also reduces as shown in Figure~\ref{fig:eps_overall}. This is because there is less energy wasted for dropping spikes.

With the help of our routing algorithm, \tech{}-TXS can still be improved upon to provide zero spike loss as shown in the results of \tech{}-SR.
\tech{}-SR achieves zero spike loss while keeping latency lower than \tech{}-SPXS and also has a similar level of energy per spike.
Hence, it can be concluded that for most spiking neuromorphic applications where spike loss is a crucial metric, we should always choose the \tech{}-SR option.

\subsection{Comparison to Other Architectures}\label{sec:archs_results}
In this section, we compare the latency, energy consumption, and energy-delay product (EDP) of four architectures: mesh-based NoC\major{~\cite{truenorth}}, butterfly NoC\major{~\cite{jiang2021butterfly}}, NeuSB\major{~\cite{balaji2022NeuSB}}, and \tech{}-SR.
\begin{figure}[h!]
	\centering
	\centerline{\includegraphics[width=1\columnwidth]{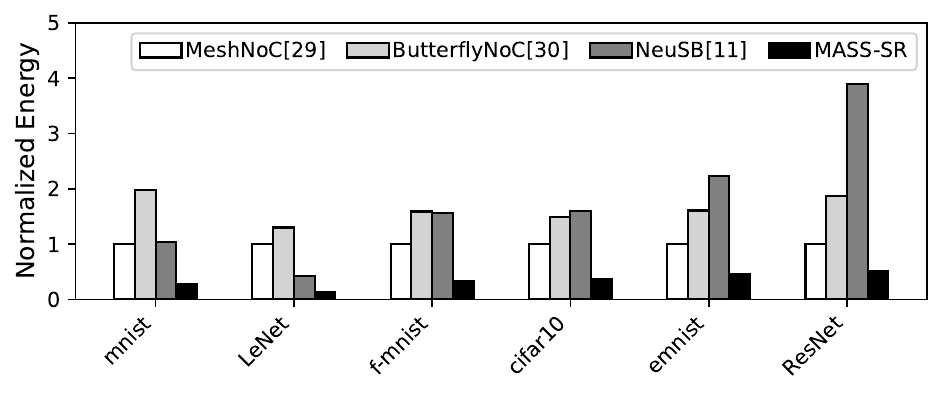}}
	\caption{\minor{Normalized energy consumption for different architectures.}}
	\label{fig:energy_comparison}
\end{figure}

The main advantage of \tech{}-SR is its energy efficiency, as shown clearly in Figure~\ref{fig:energy_comparison}. Across all applications, \tech{}-SR consistently consumes the lowest energy compare to all other architectures. On average, \tech{}-SR consumes 3.48 times less energy than Mesh NoC and 5.47 times less than Butterfly NoC. This is attributed to its lower number of switch counts and bufferless architecture.
\begin{figure}[h!]
	\centering
	\centerline{\includegraphics[width=1\columnwidth]{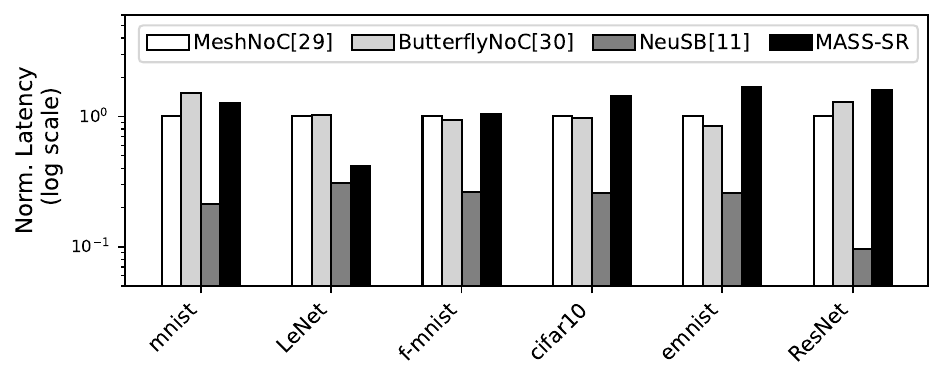}}
	\caption{\minor{Normalized latency for different architectures.}}
	\label{fig:latency_comparison}
\end{figure}

Figure~\ref{fig:latency_comparison} shows that \tech{}-SR has a slightly higher latency compared to other architectures due to spike traffic scheduling.  On average, \tech{}-SR experiences a 1.24 times increase in latency compared to Mesh NoC, with a maximum increase of 1.7 times. 
\major{When the application has a high number of spikes per connection (e.g., the LeNet benchmark), fewer scheduling delays are required, resulting in lower latency. In contrast, for applications with more uniformly distributed spike traffic, the scheduler must serialize more connections to avoid path conflicts, which naturally leads to increased latency.}
However, this latency is still negligible for the streaming application domain which we target, where the sample periods are above 1 msec. NeuSB achieves the lowest latency due to its dedicated bus lane per master, which reduces routing complexity and delays.

\begin{figure}[h!]
	\centering
	\centerline{\includegraphics[width=1\columnwidth]{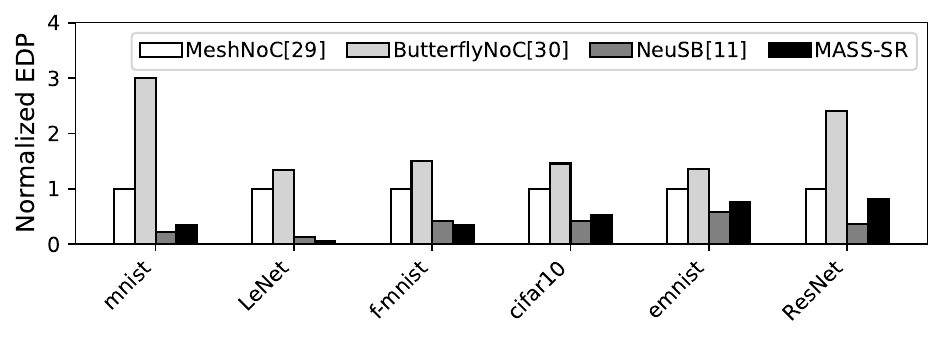}}
	\caption{\minor{Normalized EDP for different architectures.}}
	\label{fig:edp_comparison}
\end{figure}

To assess the balance between latency and energy consumption, we use the energy-delay product (EDP) as a key metric. Figure~\ref{fig:edp_comparison} illustrates normalized EDP to Mesh NoC for all compared architectures. \tech{}-SR shows a clear advantage in EDP over traditional NoCs, highlighting its efficient trade-off between latency and energy. Moreover, the design area of \tech{} is smaller than those of NoCs since there is no memory buffer. While NeuSB achieves a lower EDP than \tech{}-SR in most applications, this comes at the cost of a larger design area due to its extensive switch requirements. Given the priority of low energy consumption in neuromorphic computing, and the benefits of a compact design, \tech{}-SR remains the ideal choice for energy-efficient applications.

\section{Conclusions}\label{sec:conclusions}
In this work, we introduced \tech{}, an application mapping framework which is customized for segmented ladder bus architectures. 
Segmented ladder bus offers highly flexible interconnections, enabling any tile in the system to communicate with another through multiple routing paths. 
At runtime, this interconnect can establish simultaneous connections based on compile-time communication requirements, making it responsive to the application's dynamic needs. 
We propose a process for deploying spiking neural network applications on a dynamic segmented bus, designed to minimize spike loss and optimize energy efficiency.
This process includes three key steps: cluster mapping, traffic scheduling, and routing.
For each step, we develop heuristic algorithms to address the associated challenges effectively.

We conducted a thorough evaluation of \tech{} on ladder segmented bus in comparison with both NeuSB and traditional NoC architectures using our cycle-accurate network simulator.
Simulation results demonstrate that, when compared to conventional NoCs, \tech{} achieves 3.5 times reduction in energy consumption. Although this reduction comes with a slight increase in delay, this increment is negligible for the application domain which we target, because the speed of the bus is much faster than the communication frequency of the neuron spikes. \tech{} also offers a clear advantage in EDP over traditional NoCs. This demonstrates its efficient trade-off between latency and energy and makes its well-suited for neuromorphic computing applications.

\section*{Acknowledgment}
This work is supported by US DOE Award DE-SC0022014 and the US NSF Award CCF-1942697.

\bibliographystyle{elsarticle-num}
\bibliography{external,disco}

@inproceedings{sentryos,
  title={Design of Many-Core Big Little $\mu${Brains} for Energy-Efficient Embedded Neuromorphic Computing},
  author={Varshika, M Lakshmi and Balaji, Adarsha and Corradi, Federico and Das, Anup and Stuijt, Jan and Catthoor, Francky},
  booktitle=DATE,
  year={2022},
  optorganization={IEEE}
}

@article{spinemap,
archivePrefix = {arXiv},
arxivId = {2004.03717},
author = {Balaji, Adarsha and Das, Anup and Wu, Yuefeng and Huynh, Khanh and Dell'anna, Francesco G. and Indiveri, Giacomo and Krichmar, Jeffrey L. and Dutt, Nikil D. and Schaafsma, Siebren and Catthoor, Francky},
optdoi = {10.1109/TVLSI.2019.2951493},
opteprint = {2004.03717},
optissn = {15579999},
journal = TVLSI,
optkeywords = {Interspike interval (ISI),neuromorphic computing,spiking neural network (SNN)},
optmendeley-groups = {disco},
number = {1},
pages = {76--86},
title = {{Mapping spiking neural networks to neuromorphic hardware}},
volume = {28},
year = {2020}
}

@ARTICLE{balaji2022NeuSB,
  author={Balaji, Adarsha and Huynh, Phu Khanh and Catthoor, Francky and Dutt, Nikil D. and Krichmar, Jeffrey L. and Das, Anup},
  journal={IEEE Transactions on Emerging Topics in Computing (TETC)}, 
  title={NeuSB: A Scalable Interconnect Architecture for Spiking Neuromorphic Hardware}, 
  year={2023},
  volume={11},
  number={2},
  pages={373-387},
  doi={10.1109/TETC.2023.3238708}
}

@inproceedings{mustafazade2024clustering,
  title={Clustering and Allocation of Spiking Neural Networks on Crossbar-Based Neuromorphic Architecture},
  author={Mustafazade, Ilknur and Kandasamy, Nagarajan and Das, Anup},
  booktitle={Proceedings of the 21st ACM International Conference on Computing Frontiers},
  pages={164--171},
  year={2024}
}

@ARTICLE{huynh2024adiona,
  author={Huynh, Phu Khanh and Mustafazade, Ilknur and Catthoor, Francky and Kandasamy, Nagarajan and Das, Anup},
  journal={IEEE Embedded Systems Letters}, 
  title={A Scalable Dynamic Segmented Bus Interconnect for Neuromorphic Architectures}, 
  year={2024},
  volume={16},
  number={4},
  pages={505-508},
  doi={10.1109/LES.2024.3452551}}

@STRING{ISCA = {International Symposium on Computer Architecture (ISCA)}}

@STRING{IJCNN = {International Joint Conference on Neural Network (IJCNN)}}

@STRING{IJCNN = {International Joint Conference on Neural Networks (IJCNN)}}

@STRING{DATE = {Design, Automation \& Test in Europe Conference \& Exhibition (DATE)}}

@STRING{TVLSI = {IEEE Transactions on Very Large Scale Integration (VLSI) Systems}}

@STRING{ASAP = {International Conference on Application-specific Systems, Architectures, and Processors (ASAP)}}

@article{truenorth,
  title={{TrueNorth: Accelerating from zero to 64 million neurons in 10 years}},
  author={DeBole, Michael V and Taba, Brian and Amir, Arnon and others},
  optauthor={DeBole, Michael V and Taba, Brian and Amir, Arnon and Akopyan, Filipp and Andreopoulos, Alexander and Risk, William P and Kusnitz, Jeff and Otero, Carlos Ortega and Nayak, Tapan K and Appuswamy, Rathinakumar and others},
  journal={Computer},
  optvolume={52},
  optnumber={5},
  optpages={20--29},
  year={2019},
  optpublisher={IEEE}
}

@inproceedings{carlsim6,
  title={CARLsim 6: an open source library for large-scale, biologically detailed spiking neural network simulation},
  author={Niedermeier, Lars and Chen, Kexin and Xing, Jinwei and others},
  optauthor={Niedermeier, Lars and Chen, Kexin and Xing, Jinwei and Das, Anup and Kopsick, Jeffrey and Scott, Eric and Sutton, Nate and Weber, Killian and Dutt, Nikil and Krichmar, Jeffrey L},
  booktitle={2022 International Joint Conference on Neural Networks (IJCNN)},
  pages={1--10},
  year={2022},
  organization={IEEE}
}

@article{rathi2023exploring,
  title={Exploring neuromorphic computing based on spiking neural networks: Algorithms to hardware},
  author={Rathi, Nitin and Chakraborty, Indranil and Kosta, Adarsh and Sengupta, Abhronil and Ankit, Aayush and Panda, Priyadarshini and Roy, Kaushik},
  journal={ACM Computing Surveys},
  optvolume={55},
  optnumber={12},
  optpages={1--49},
  year={2023},
  optpublisher={ACM New York, NY}
}

@article{modha2023neural,
  title={Neural inference at the frontier of energy, space, and time},
  author={Modha, Dharmendra S and Akopyan, Filipp and Andreopoulos, Alexander and others},
  optauthor={Modha, Dharmendra S and Akopyan, Filipp and Andreopoulos, Alexander and Appuswamy, Rathinakumar and Arthur, John V and Cassidy, Andrew S and Datta, Pallab and DeBole, Michael V and Esser, Steven K and Otero, Carlos Ortega and others},
  journal={Science},
  optvolume={382},
  optnumber={6668},
  optpages={329--335},
  year={2023},
  publisher={American Association for the Advancement of Science}
}

@inproceedings{jin2023mapping,
  title={Mapping very large scale spiking neuron network to neuromorphic hardware},
  author={Jin, Ouwen and Xing, Qinghui and Li, Ying and Deng, Shuiguang and He, Shuibing and Pan, Gang},
  booktitle={Proceedings of the 28th ACM International Conference on Architectural Support for Programming Languages and Operating Systems, Volume 3},
  pages={419--432},
  year={2023}
}

@article{neumap,
  title={Optimal mapping of spiking neural network to neuromorphic hardware for edge-AI},
  author={Xiao, Chao and Chen, Jihua and Wang, Lei},
  journal={Sensors},
  volume={22},
  number={19},
  pages={7248},
  year={2022},
  publisher={MDPI}
}

@article{richter2024dynap,
  title={DYNAP-SE2: a scalable multi-core dynamic neuromorphic asynchronous spiking neural network processor},
  author={Richter, Ole and others},  
  opauthor={Richter, Ole and Wu, Chenxi and Whatley, Adrian M and K{\"o}stinger, German and Nielsen, Carsten and Qiao, Ning and Indiveri, Giacomo},
  journal={Neuromorphic Computing and Engineering},
  volume={4},
  number={1},
  pages={014003},
  year={2024},
  publisher={IOP Publishing}
}

@article{xiao2017fashion,
  title={Fashion-mnist: a novel image dataset for benchmarking machine learning algorithms},
  author={Xiao, Han and Rasul, Kashif and Vollgraf, Roland},
  journal={arXiv preprint arXiv:1708.07747},
  year={2017}
}

@inproceedings{cohen2017emnist,
  title={EMNIST: Extending MNIST to handwritten letters},
  author={Cohen, Gregory and Afshar, Saeed and Tapson, Jonathan and Van Schaik, Andre},
  booktitle={2017 international joint conference on neural networks (IJCNN)},
  pages={2921--2926},
  year={2017},
  organization={IEEE}
}

@ARTICLE{snntorch,
  author={Eshraghian, Jason K. and Ward, Max and Neftci, Emre O. and Wang, Xinxin and Lenz, Gregor and Dwivedi, Girish and Bennamoun, Mohammed and Jeong, Doo Seok and Lu, Wei D.},
  journal={Proceedings of the IEEE}, 
  title={Training Spiking Neural Networks Using Lessons From Deep Learning}, 
  year={2023},
  volume={111},
  number={9},
  pages={1016-1054},
  doi={10.1109/JPROC.2023.3308088}}

@article{ekici2021bin,
  title={Bin packing problem with conflicts and item fragmentation},
  author={Ekici, Ali},
  journal={Computers \& Operations Research},
  volume={126},
  pages={105113},
  year={2021},
  publisher={Elsevier}
}

@misc{gary1979computers,
  title={Computers and Intractability: A Guide to the Theory of NP-completeness},
  author={Gary, Michael R and Johnson, David S},
  year={1979},
  publisher={WH Freeman and Company, New York}
}

@article{selman2006hill,
  title={Hill-climbing search},
  author={Selman, Bart and Gomes, Carla P},
  journal={Encyclopedia of cognitive science},
  volume={81},
  number={333-335},
  pages={10},
  year={2006},
  publisher={Wiley New York, NY, USA}
}

@inproceedings{he2016deep,
  title={Deep residual learning for image recognition},
  author={He, Kaiming and Zhang, Xiangyu and Ren, Shaoqing and Sun, Jian},
  booktitle={Proceedings of the IEEE conference on computer vision and pattern recognition},
  pages={770--778},
  year={2016}
}

@article{wolfe2010sparse,
  title={Sparse and powerful cortical spikes},
  author={Wolfe, Jason and Houweling, Arthur R and Brecht, Michael},
  journal={Current opinion in neurobiology},
  volume={20},
  number={3},
  pages={306--312},
  year={2010},
  publisher={Elsevier}
}

@article{olshausen2004sparse,
  title={Sparse coding of sensory inputs},
  author={Olshausen, Bruno A and Field, David J},
  journal={Current opinion in neurobiology},
  volume={14},
  number={4},
  pages={481--487},
  year={2004},
  publisher={Elsevier}
}

@inproceedings{li2020sneap,
  title={SNEAP: A fast and efficient toolchain for mapping large-scale spiking neural network onto NoC-based neuromorphic platform},
  author={Li, Shiming and Guo, Shasha and Zhang, Limeng and Kang, Ziyang and Wang, Shiying and Shi, Wei and Wang, Lei and Xu, Weixia},
  booktitle={Proceedings of the 2020 on Great Lakes Symposium on VLSI},
  pages={9--14},
  year={2020}
}

@article{raghavan2008distributed,
  title={Distributed loop controller for multithreading in unithreaded ILP architectures},
  author={Raghavan, Praveen and Lambrechts, Andy and Jayapala, Murali and Catthoor, Francky and Verkest, Diederik},
  journal={IEEE Transactions on Computers},
  volume={58},
  number={3},
  pages={311--321},
  year={2008},
  publisher={IEEE}
}

@article{duchovn2014path,
  title={Path planning with modified a star algorithm for a mobile robot},
  author={Ducho{\v{n}}, Franti{\v{s}}ek and Babinec, Andrej and Kajan, Martin and Be{\v{n}}o, Peter and Florek, Martin and Fico, Tom{\'a}{\v{s}} and Juri{\v{s}}ica, Ladislav},
  journal={Procedia engineering},
  volume={96},
  pages={59--69},
  year={2014},
  publisher={Elsevier}
}

@inproceedings{moscibroda2009case,
  title={A case for bufferless routing in on-chip networks},
  author={Moscibroda, Thomas and Mutlu, Onur},
  booktitle={ISCA},
  optpages={196--207},
  year={2009}
}

@inproceedings{michelogiannakis2010evaluating,
  title={Evaluating bufferless flow control for on-chip networks},
  author={Michelogiannakis, George and others},  
  opauthor={Michelogiannakis, George and Sanchez, Daniel and Dally, William J and Kozyrakis, Christos},
  booktitle={NOCS},
  opbooktitle={2010 Fourth ACM/IEEE International Symposium on Networks-on-Chip},
  oppages={9--16},
  year={2010},
  oporganization={IEEE}
}

@inproceedings{catania2015noxim,
  title={Noxim: An open, extensible and cycle-accurate network on chip simulator},
  author={Catania, Vincenzo and Mineo, Andrea and others}, 
  opauthor={Catania, Vincenzo and Mineo, Andrea and Monteleone, Salvatore and Palesi, Maurizio and Patti, Davide},
  booktitle={ASAP},
  oppages={162--163},
  year={2015},
  oporganization={IEEE}
}

@ARTICLE{systemC,
  author={},
  journal={IEEE Std 1666-2023 (Revision of IEEE Std 1666-2011)}, 
  title={IEEE Standard for Standard SystemC® Language Reference Manual}, 
  year={2023},
  volume={},
  number={},
  pages={1-618},
  doi={10.1109/IEEESTD.2023.10246125}
}

@inproceedings{jiang2021butterfly,
  title={Design of Multi-core Spiking Neural Network Chip Based on Butterfly Network},
  author={Jiang, Hao and Wei, Jinsong and Li, Ye and Lu, Jikai and Shi, Tuo and Liu, Qi},
  booktitle={2021 IEEE 14th International Conference on ASIC (ASICON)},
  pages={1--4},
  year={2021},
  organization={IEEE}
}

@article{serrano2013stdp,
  title={STDP and STDP variations with memristors for spiking neuromorphic learning systems},
  author={Serrano-Gotarredona, Teresa and Masquelier, Timoth{\'e}e and Prodromakis, Themistoklis and Indiveri, Giacomo and Linares-Barranco, Bernabe},
  journal={Frontiers in neuroscience},
  volume={7},
  pages={2},
  year={2013},
  publisher={Frontiers Media SA}
}

\vspace{1em}
\Needspace{8\baselineskip}
\setlength\intextsep{0pt}
\begin{wrapfigure}{l}{0.25\columnwidth} 
    \includegraphics[width=1in,height=1.25in,clip,keepaspectratio]{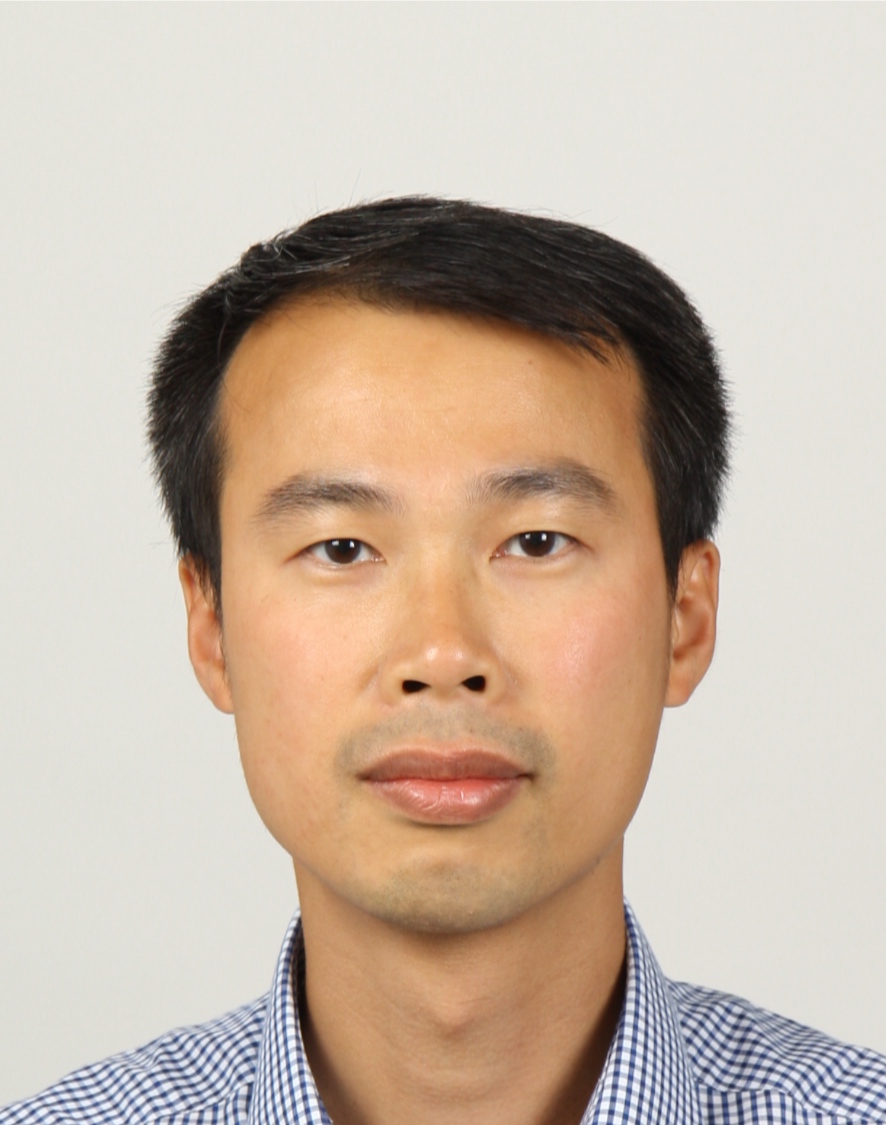}
\end{wrapfigure}
\major{\noindent\textbf{Phu Khanh Huynh} received a Ph.D. degree from the Department of Electrical and Computer Engineering, Drexel University, Philadelphia, PA in 2025. His research interests include designing, exploration, and optimization of neuromorphic computing systems, with focus on spiking neural networks (SNN).}

\vspace{1em}
\Needspace{8\baselineskip}
\setlength\intextsep{0pt}
\begin{wrapfigure}{l}{0.25\columnwidth} 
    \includegraphics[width=1in,height=1.25in,clip,keepaspectratio]{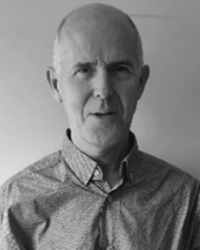}
\end{wrapfigure}
\major{\noindent\textbf{Francky Catthoor} received the engineering degree and a Ph.D. in electrical engineering from the Katholieke Universiteit Leuven, Belgium, in 1982 and 1987 respectively. Between 1987 and 2000, he headed several research domains in the area of high-level and system synthesis techniques and architectural methodologies at IMEC. Since 2000 he has also been strongly involved in other activities including co-exploration of application, computer architecture and deep submicron technology aspects, biomedical systems and IoT sensor nodes, and photovoltaic modules combined with renewable energy systems, all at the Inter-university Micro-Electronics Center (IMEC), Heverlee, Belgium. After his official retirement from IMEC in Oct 2024, he is now affiliated with the Microlab of NTUAthens in Greece as an associated guest professor. He is emeritus full professor at the EE department of the K.U.Leuven. He was elected an IEEE Fellow in 2005.}

\vspace{1em}
\Needspace{8\baselineskip}
\setlength\intextsep{0pt}
\begin{wrapfigure}[8]{l}{0.25\columnwidth} 
    \includegraphics[width=1in,height=1.25in,clip,keepaspectratio]{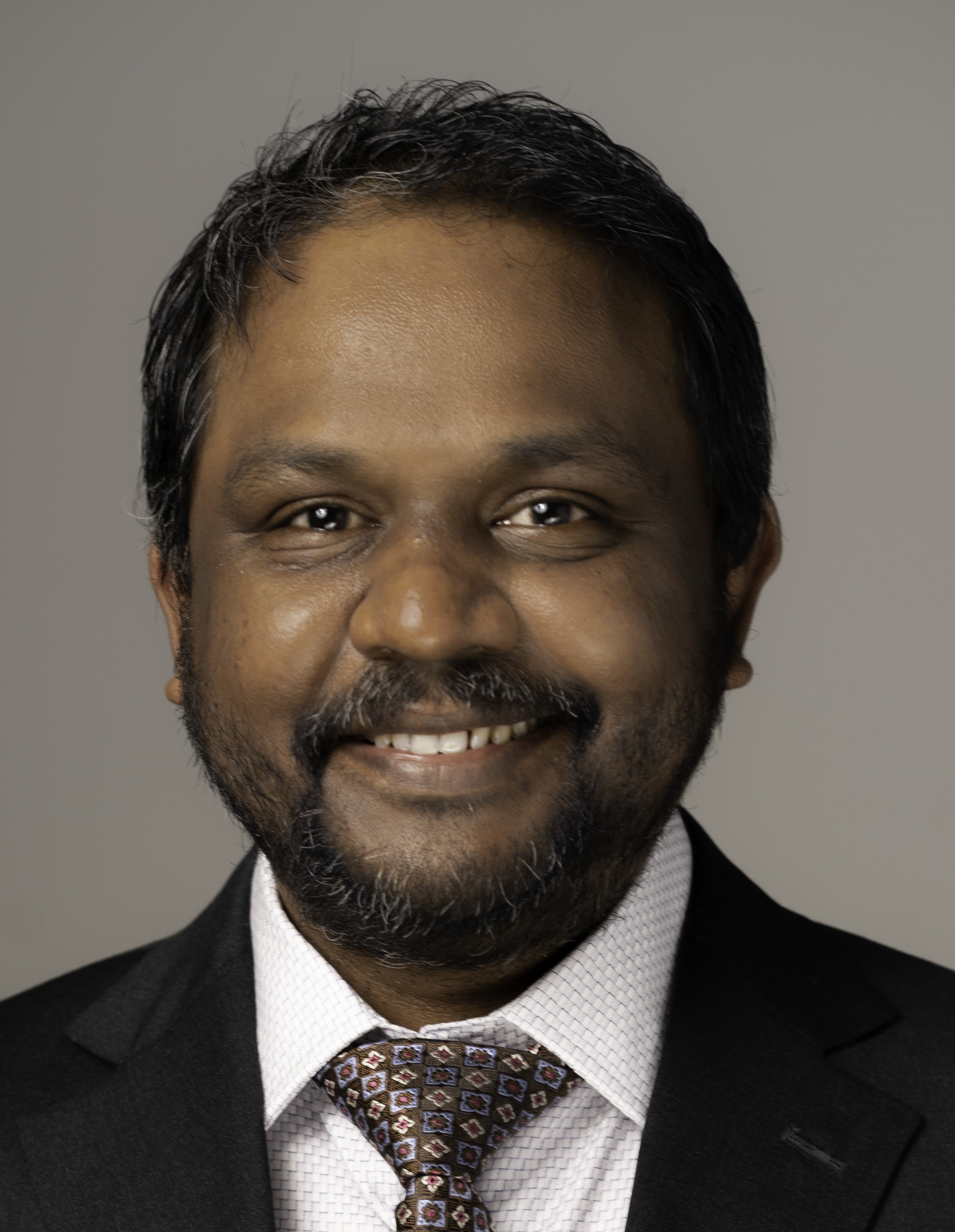}
\end{wrapfigure}
\major{\noindent\textbf{Anup Das} is an Associate Professor at Drexel University and Associate Department Head for Graduate Studies. He received a Ph.D. in Embedded Systems from the National University of Singapore in 2014.  Following his Ph.D., he was a postdoctoral fellow at the University of Southampton and a researcher at IMEC. His research focuses on neuromorphic computing and architectural exploration. 
He received the United States National Science Foundation CAREER Award in 2020 and the Department of Energy CAREER Award in 2021 to investigate the reliability and security of neuromorphic hardware.
He is a senior member of the IEEE and a member of the ACM.}

\end{document}